\documentclass[runningheads]{llncs}
\usepackage{graphicx}
\usepackage{comment}
\usepackage{amsmath,amssymb} 
\usepackage{color}

\usepackage[colorlinks]{hyperref}
\hypersetup{final}

\newcommand{\enc}{\textrm{enc}}
\newcommand{\dec}{\textrm{dec}}

\newcommand{\V}{\mathcal{V}}
\newcommand{\E}{\mathcal{E}}

\begin{document}
\pagestyle{headings}
\mainmatter
\def\ECCVSubNumber{}  

\title{Intrinsic Point Cloud Interpolation via \\
Dual Latent Space Navigation} 

\titlerunning{Intrinsic Point Cloud Interpolation}
%
\author{Marie-Julie Rakotosaona\inst{1} \and
Maks Ovsjanikov\inst{1}}
\authorrunning{Rakotosaona M-J., Ovsjanikov M.}
%
\institute{LIX, Ecole Polytechnique \\
\email{\{mrakotos,maks\}@lix.polytechnique.fr}}
\maketitle

\begin{abstract}
 We present a learning-based method for interpolating and manipulating 3D shapes represented as point clouds, that is
  explicitly designed to preserve intrinsic shape properties. Our approach is based on constructing a dual encoding
  space that enables shape synthesis and, at the same time, provides
  links to the intrinsic shape information, which is
  typically not available on point cloud data. Our method works in a single pass and avoids expensive optimization,
  employed by existing techniques. Furthermore, the strong regularization provided by our dual latent space approach
  also helps to improve shape recovery in challenging settings from noisy point clouds across different datasets. Extensive
  experiments show that our method results in more realistic and smoother interpolations compared to baselines.

\end{abstract}

\section{Introduction}
A core problem in 3D computer vision is to analyze, encode and manipulate shapes represented as point clouds. Point clouds are particularly useful compared to other representations due to their generality, simplicity and flexibility compared to more complex data-structures such as triangle meshes or dense voxel grids. For all of these reasons, and with the introduction of PointNet and its variants \cite{qi2017pointnet,qi2017pointnet++,shen2018mining}, point clouds have also gained popularity in machine learning applications, including point-based \emph{generative models}.

\begin{figure}[t]
\begin{center}
   \includegraphics[width=0.9\linewidth]{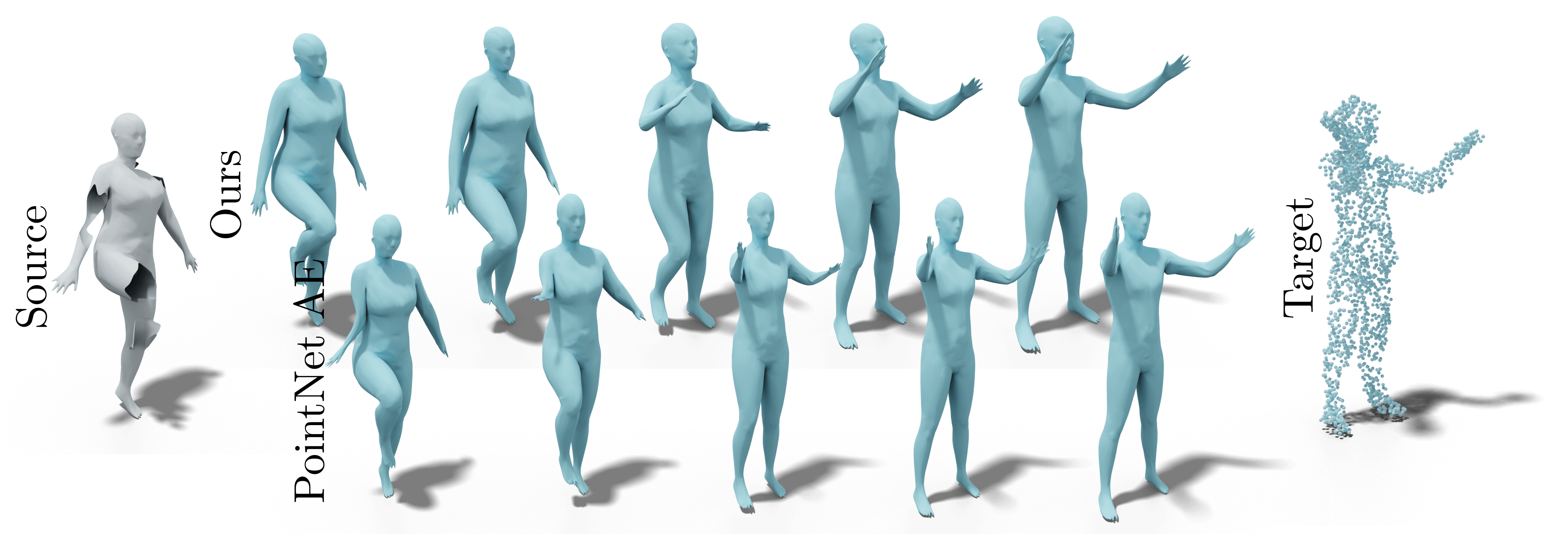}
\end{center}
   \caption{Intrinsic point cloud interpolation between points from an incomplete scan with holes  (left, reconstructed in first blue column) and  points from a noisy mesh (right, reconstructed in last blue column). Our method both reconstructs the shape better and produces a more natural interpolation than a PointNet-based auto-encoder}
\label{fig:teaser}

\end{figure}

Unfortunately the flexibility of point cloud representations also comes at a cost, as they do not encode any topological or intrinsic metric information of the underlying surface. Thus, methods trained on point cloud data can by their nature be insensitive to distortion that might appear on generated shapes. This problem is particularly prominent in 3D shape interpolation, where a common approach is to generate intermediate shapes by interpolating the learned latent vectors. In this case, even if the end-shapes are realistic, the intermediate ones can have severe distortions that are very difficult to detect and correct using only point-based information. More generally, several works have observed that generative models built on point cloud data can fail to capture the space of natural shapes, e.g., \cite{li2018point,huang2019operatornet}, making it difficult to navigate them while maintaining realism.

In this paper, we introduce a novel architecture aimed specifically at injecting intrinsic information into a generative
point-based network. Our method works by learning consistent mappings across the latent space obtained by a point cloud auto-encoder
 and another feature encoding that captures the intrinsic shape structure. We show that these two parts
can be optimized jointly using shapes represented as triangle meshes during training. The resulting linked latent space
combines the strengths of a generative latent model, and the intrinsic surface information. Finally, we use
the learned networks at test time on raw 3D point clouds that are neither in correspondence with the training shapes,
nor contain any connectivity information. 

Our approach is general and not only leads to smooth interpolations, while avoiding expensive iterative optimization, but also, as we show bellow, leads to more accurate shape reconstruction from noisy point clouds across different datasets. We demonstrate on a wide range of experiments that our approach can significantly
improve upon recent baselines in terms of the accuracy and smoothness of the interpolation and enables a range of novel applications.

\section{Related Work}
\label{sec:related}

Shape interpolation, also known as morphing in certain contexts, and exploration is a vast and well-researched area of
computer vision and computer graphics (see \cite{lazarus1998three} for a survey of the early approaches) and its full
overview is beyond the scope of this paper. Below we review works most closely related to ours, and concentrate, in
particular, on either structure-preserving mesh interpolation techniques, or recent learning-based methods that focus on
point clouds.

Classical methods for 3D shape interpolation have primarily focused on designing well-founded geometric metrics, and
associated optimization methods that enable smooth structure-preserving interpolations. Early works in this direction
include variants of as-rigid-as-possible interpolation and modeling
\cite{alexa2000rigid,igarashi2005rigid,xu2006poisson} and various \emph{representations} of shape deformation that
facilitate specific transformation types,
e.g. \cite{von2006vector,huang2006subspace,lipman2007volume,crane2011spin,vaxman2015conformal} among many others.

A somewhat more principled framework is provided by the notion of \emph{shape spaces}
\cite{kendall1984shape,michor2006riemannian} in which interpolation can be phrased as computing a shortest path
(geodesic). In the case of surface meshes, this approach was studied in detail in \cite{kilian2007geometric} and then
extended in numerous follow-up works, including
\cite{wirth2011continuum,freifeld2012lie,heeren2012time,heeren2014exploring,heeren2016splines} among many others. These approaches enjoy a
rich theoretical foundation, but are typically restricted to shapes having a fixed connectivity and can lead to
difficult optimization problems at test time.

We also note a recent set of methods based on the formalism of \emph{optimal transport}
\cite{benamou2000computational,solomon2015convolutional,bonneel2015sliced} which have also been used for shape
interpolation. These approaches treat the input shapes as probability measures that are interpolated via efficient
optimization techniques. 

Somewhat more closely related to ours are data-driven and feature-based interpolation methods. These include
interpolation based on hand-crafted features \cite{gao2017data,huang2019operatornet} or by exploring various
\emph{local} shape spaces obtained by analyzing a shape collection
\cite{gao2013data,zhang2015fast,von2016optimized}. These approaches work well if the input shapes are sufficiently
similar, but require triangle meshes and dense point-wise correspondences, or a single template that is fitted to all
input data to build a statistical model, e.g. \cite{hasler2009statistical,bogo2014faust,bogo2017dynamic}.

Most closely related to ours are recent generative models that operate directly on point clouds
\cite{achlioptas2018learning,li2018point,liu2019l2g}. These methods are largely inspired by the seminal work
of PointNet and its variants \cite{qi2017pointnet,qi2017pointnet++} and are typically based on autoencoder architectures
that allow shape exploration by manipulation in the latent space. Despite significant
progress in this area, however, the structure of learned latent spaces is typically not easy to control or analyze. For
example, it is well-known (see e.g.  \cite{huang2019operatornet}) that commonly used linear interpolation in latent
space can give rise to unrealistic shapes that are difficult to detect and rectify. 

Common approaches to address these issues include extensive data augmentation \cite{groueix20183d}, using adversarial
losses that aim to penalize unrealistic instances \cite{li2018point,ben2018multi} or modifying the metric in the latent
space. The latter can be done by computing the Jacobian of the decoder from the latent to the embedding space
\cite{chen2017metrics,shao2018riemannian} or using feature-based metrics at test time
\cite{laine2018feature,frenzel2019latent}. Unfortunately, as we show below such techniques either lead to difficult
optimization problems at test time, or can still result in significant shape distortion.

\vspace{-2mm}
\subsubsection{Contribution} In this paper, we propose to address this challenge by building a \emph{dual latent space}
that combines a learned shape encoding in a point-based generative model with another parallel encoding that aims to
capture the intrinsic shape metric given by the lengths of edges of triangle meshes only required during training. This
second encoding exploits the insights of mesh-based interpolation techniques
\cite{kilian2007geometric,heeren2016splines,sassen2019solving} that highlight the importance of \emph{interpolating the
  intrinsic surface information} rather than the point coordinates. We combine these two encodings by constructing dense networks that 
``translate'' between the two latent spaces, and enable smooth and accurate interpolation at test time without relying on
correspondences or expensive optimization problems.

\section{Motivation \& Background}
Our main goal is to design a method capable of \emph{efficiently and accurately} interpolating shapes represented as point clouds. This problem is challenging for several key reasons. First, most existing theoretically well-founded axiomatic 3D shape interpolation methods \cite{kilian2007geometric,heeren2012time,heeren2014exploring,heeren2016splines} assume the input shapes to be represented as triangle meshes with fixed connectivity in 1-1 correspondence, and furthermore typically require extensive optimization at test time. On the other hand, learning-based approaches typically embed the shapes in a compact latent space, and interpolate shapes by linearly interpolating their corresponding latent vectors \cite{achlioptas2018learning,wu2016learning}. Although this approach is efficient, the metric in the latent space is typically not well-understood and therefore \emph{linear interpolation} in this space may result in unrealistic and heavily distorted shapes. Classical methods such as Variational Auto-Encoders (VAEs) help introduce regularity into the latent space, and enable more accurate generative models, but offer little control on the distances and thus interpolation in the latent space. To address this challenge, several recent approaches have proposed ways to endow the latent space with a metric and help recover geodesic distances \cite{laine2018feature,chen2017metrics,frenzel2019latent}. However, these methods again typically involve expensive computations such as the Jacobian of the decoder network, and expensive \emph{optimization at test time}.

Within this context, our main goal is to combine the formalism and shape metrics proposed by geometric methods \cite{kilian2007geometric,heeren2016splines} with the accuracy and flexibility of data-driven techniques while maintaining efficiency and scalability.

\subsubsection{Shape Interpolation Energy}
We first recall the intrinsic shape interpolation energy introduced in \cite{kilian2007geometric}. Specifically suppose we are given a pair of shapes $M,N$ represented as triangle meshes with fixed connectivity, so that $M=(\V_M, \E)$, and $N=(\V_N, \E)$, where $\V, \E$ represent the coordinates of the points and the fixed set of edges respectively. An interpolating sequence is defined by a one parameter family $S_t = (\V_t, \E)$, such that $\V_0 = \V_M$, and $\V_1 = \V_N$. Denoting by $v_i(t)$ the trajectory of vertex $i$ in $S_t$, the basic time-continuous intrinsic interpolation energy of $S_t$ is defined as:
\begin{align}
  E_{\text{cont}}(S_t) = \int_{t=0}^1 \sum_{(i,j) \in \mathcal{E}} \left(\frac{\partial \|v_i(t) - v_j(t) \|_2}{\partial t}\right)^2 dt.
\end{align}

This energy measures the integral of the change of all the edge lengths in the interpolation sequence. It can be discretized in time by sampling the interval $[0 \ldots 1]$ with samples $t_k$, where $k=1 \ldots n_k$. When these time samples are uniform, resulting in a discrete set of shapes $\{S_k\}$, this leads to the discrete energy:
\begin{equation}
\label{eq:disc_energy}
\begin{aligned}
  E_{\text{disc}}(\{S_k\}) = \sum_{k=2}^{n_k} \sum_{ij \in \mathcal{E}} \left(\|v_i(t_k) -v_j(t_k) \|_2 - \|v_i(t_{k-1}) - v_j(t_{k-1}) \|_2 \right)^2.
\end{aligned}
\end{equation}
This discrete energy simply measures the sum of the squared differences between lengths of edges across consecutive shapes in the sequence. The authors of \cite{kilian2007geometric} argue that computing a shape sequence between $M$ and $N$ that minimizes such a distortion energy results in an accurate interpolation of the two shapes (more precisely in \cite{kilian2007geometric} an additional weak regularization is employed, which we omit for simplicity and as we have found it to be unnecessary in our case).  Note that both the continuous and discrete versions of the energy promote \emph{as-isometric-as-possible} shape interpolations. Specifically they aim to minimize the \emph{isometric distortion} by promoting intermediate meshes whose edge lengths \emph{interpolate as well as possible} the edge lengths of $M, N$, without requiring the two input shapes to be isometric themselves.

Despite the simplicity and elegance of the intrinsic interpolation energy, minimizing it directly is challenging as it leads to large non-convex optimization problems over vertex coordinates. Indeed, additional regularization is typically required to achieve realistic interpolation across large motions \cite{kilian2007geometric,heeren2016splines}. Perhaps even more importantly, the assumption of input shapes having a fixed triangle mesh and being in 1-1 correspondence is very restrictive in practice.

\subsubsection{Latent space optimization}
In the context of data-driven techniques the standard way to manipulate shapes is through operations in the \emph{latent space}, by first training an auto-encoder (AE) architecture and then shape manipulation (e.g. interpolation) in the learned latent space. Specifically, an encoder is trained to associate a \emph{latent vector} $l_S$ to each 3D shape $S$ in a training set via $l_S = \enc(S)$, while the decoder is trained so that $\dec(l_S) \approx S$. Given two shapes $M, N$, the interpolation is done by first computing their latent vectors, $l_{M}, l_{N}$ and then constructing an interpolating sequence via $S_{t} = \dec ( t l_{N} + (1-t) l_M)$ \cite{achlioptas2018learning,wu2016learning}.

Unfortunately, basic \emph{linear interpolation} in the latent space can produce significant artefacts in the resulting reconstructed shapes as we can see in Figure \ref{fig:example_interp}. More broadly, the metric  (distance) structure of the latent space is not easy to control, as the encoder-decoder architecture is typically trained only to be able to \emph{reconstruct} the shapes, and does not capture any information about distances in the latent space. 


\subsection{Metric interpolation in a learned space}
To overcome this limitation, perhaps the simplest approach is to use a learned latent space, but to compute an interpolating sequence while minimizing the intrinsic distortion energy of the decoded shapes explicitly. 

Namely, after training an auto-encoder, given the source and target shapes with latent vectors $l_M, l_N$, one can construct a set of samples $l_k$ in the latent space  and \emph{at test time} optimize:
\begin{equation}
\label{eq:disc_opt}
\begin{aligned}
\min_{l_1, l_2, \ldots, l_k} E_{\text{disc}}(\{S_k\}), {\rm{s. t. } } ~  S_{i} = \dec(l_i),  i=1 \ldots k, \\
S_0 = \dec (l_M), S_{k+1} = \dec (l_N).
\end{aligned} 
\end{equation}
This operation employs the fact that a decoder can be trained to always produce shapes that are in 1-1 correspondence, thus making it possible to compare the decoded shapes $\{S_k\}$.

To solve this problem, the samples $l_k$ can be initialized through linear interpolation of $l_M, l_N$, and Eq.~\eqref{eq:disc_opt} can optimized via gradient descent using the pre-trained decoder network. This is significantly more efficient than directly optimizing Eq.~\eqref{eq:disc_energy} through the coordinates of the vertices, as the dimensionality of the latent space is typically much smaller. Intuitively, this procedure locally adjusts the latent vectors to correct the distortion induced by using the Euclidean metric in the latent space. In addition, the use of a pre-trained decoder acts as the regularization (required by purely geometric methods) to produce realistic shapes.

Despite leading to significant improvement compared to the basic linear interpolation in the latent space, this approach has two key limitations 1)  it requires potentially expensive optimization at test time, and 2) its accuracy is limited by the initial linear interpolation in the latent space. The latter issue is particularly prominent since the latent space is not related to the intrinsic distortion energy and therefore linear interpolation can be a suboptimal initialization for the problem in Eq.~\eqref{eq:disc_opt}.

\subsubsection{Intuition}

To design our approach we propose to build two auto-encoder networks: one that intuitively creates a parametrization of the set of realistic shapes, and the other that captures intrinsic distortion, and thus distances between shapes in shape space. This second network builds a latent space that encodes lengths of edges of underlying meshes (available at training) so that Euclidean distances in the latent space correspond to distances between lists of ordered edges. Our main intuition is that in the absence of any constraints the intrinsic distortion energy $E_{\text{disc}}$ is minimized by the family of shapes that linearly interpolates the edge lengths between the source and the target. This, however is not guaranteed to lead to actual 3D shapes, both because additional integrability conditions must hold to ensure that edges can be assembled into a consistent mesh \cite{wang2012linear} and because interpolated shapes might not be realistic from the point of view of the training data. Therefore, we also build two ``translation'' or mapping networks that allow us to go between the edge length and shape latent spaces. Finally, after training these networks, at test time, we linearly interpolate in the edge length latent space, but recover each shape by projecting onto the shape space and reconstructing using the shape decoder. As we show below, this results in both smooth and realistic shape interpolation, without relying on correspondences or optimization at test time.

\section{Method}
\subsection{Overview}

\begin{figure}
  \vspace{-10mm}
\centering
\begin{minipage}{.5\textwidth}
  \vspace{5mm}
  \includegraphics[width=1\linewidth]{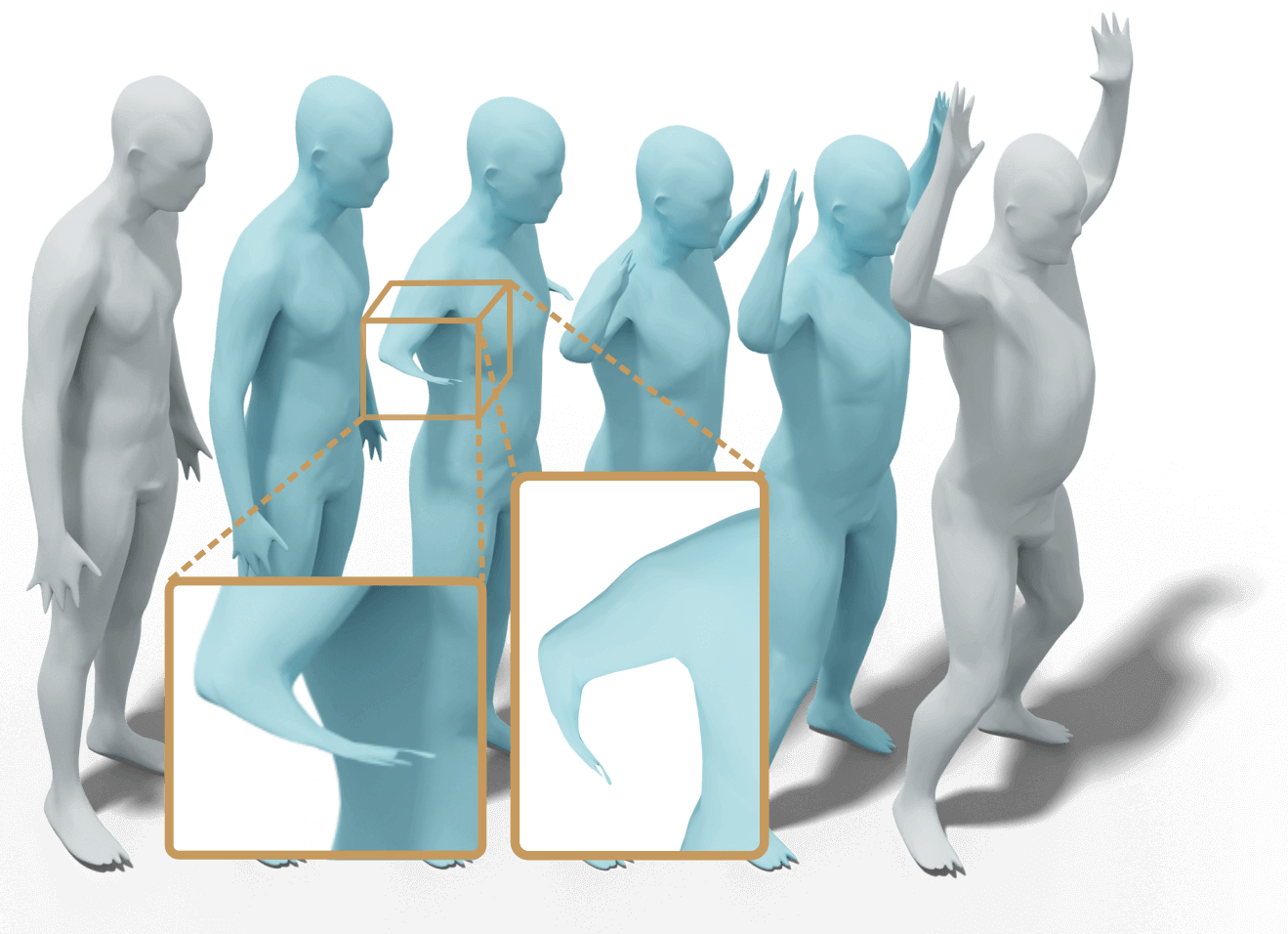}
  \caption{Linear interpolation in the latent space of the shape AE produces artefacts, as the interpolation is close to linear interpolation of the coordinates.}
  \label{fig:example_interp}
\end{minipage}%
\hfill
\begin{minipage}{.45\textwidth}
  \includegraphics[width=1\linewidth]{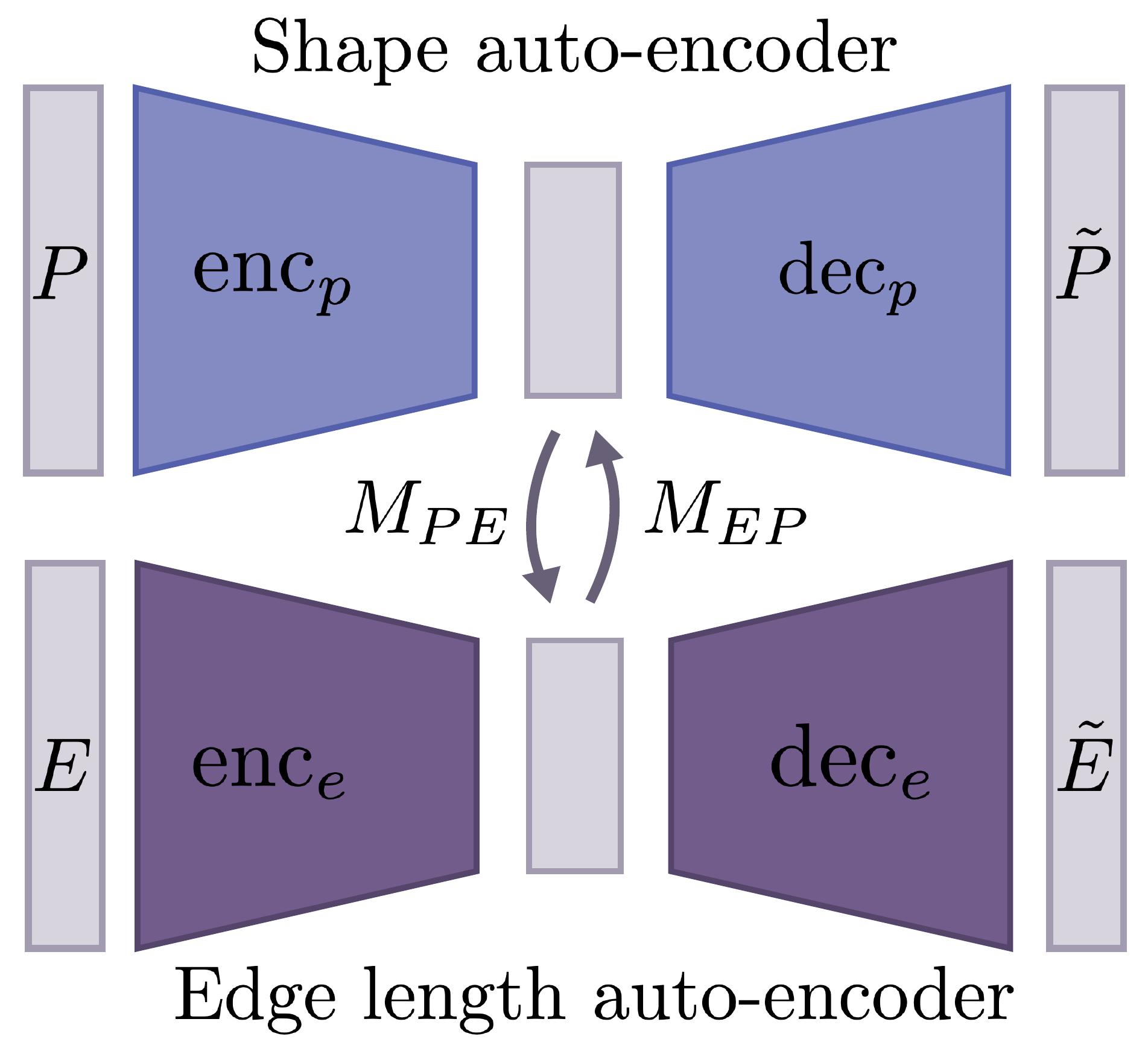}
  \caption{Our overall architecture. We build two auto-encoders that capture the shape and edge length structure respectively, as well as two mapping networks $M_{PE}$ and $M_{EP}$ that ``translate'' across the two latent spaces.\vspace{-10mm}}
  \label{fig:architecture}
\end{minipage}
\end{figure}

Figure~\ref{fig:architecture} gives an overview of our network.  As mentioned above, it consists of three main building blocks and training steps: a Shape auto-encoder, an auto-encoder of the edge lengths of the underlying mesh, and two ``translation'' networks that enable communication between the two latent spaces. These networks are used at test time to endow given point clouds with intrinsic information which is then used, in particular, for more accurate point cloud interpolation. We assume that the training data is given in the form of triangle meshes with fixed connectivity, while the input at test time consists of unorganized point clouds. In the following section we describe our architecture and the associated losses, while the implementation and experimental details are given in Section \ref{sec:results}.




\vspace{-2mm}
\subsection{Architecture} \label{sec:architecture}


\subsubsection{Shape auto-encoder.} Our first building block (Figure \ref{fig:architecture} top) consists of a shape auto-encoder, based on the PointNet architecture \cite{qi2017pointnet}. We denote the encoder and decoder networks as $\enc_p$ and $\dec_p$ respectively (we provide the exact implementation details \ref{sec_supp:architecture}).  To train this network we use the basic $L_2$ reconstruction loss, since we assume that the input shapes are in 1-1 correspondence. This leads to the following training loss:
%
\begin{equation}\label{equ:shape_ae_loss}
  \vspace{-1.5mm}
  L_{rec}(P) =\frac {1}{n}\sum_{i=1}^{n}\| P_{i}-{{\tilde{P}_{i}}}\|^{2}, \text{ where } \tilde{P} = \dec_p \left( \enc_p (P) \right).
  \vspace{-1.5mm}
 \end{equation}
Here $P$ is a training shape, the summation is done over all points in the point cloud, and $P_i$ represents the 3D coordinates of point $i$.

Importantly, our point-based encoder $\enc_{p}$ inherits the permutation invariance of PointNet \cite{qi2017pointnet}, which is crucial in real applications. Specifically, this allows us to encode arbitrary point clouds at test time even if they have significantly different sampling and are not in correspondence with the training data.

\vspace{-4mm}
 \subsubsection{Edge length auto-encoder} As observed in previous works and as we confirm bellow, the shape AE can capture the structure of individual shapes, but often fails to reflect the overall structure of \emph{shape space}, which is particularly evident in shape interpolation applications. We address this issue by constructing a separate auto-encoder that aims to capture the intrinsic shape information, and by learning mappings across the two latent spaces.

For this, we first build an auto-encoder ($\enc_e$, $\dec_e$) with dense layers that aims to reconstruct a list of edge lengths. Note that since we assume 1-1 correspondence at training time, the list of lengths of edges can be given in canonical (e.g., lexicographic with respect to vertex ids) order. We therefore build an auto-encoder that encodes this list into a compact vector and decodes it back from the latent representation. Our training loss for this part consists of two components: an $L_2$ error on the predicted edge lengths and an additional term that promotes linearity in the learned latent space:
\begin{align}\label{equ:lin_latent_space1}
L_{e}(E_A) &= \|\dec_e(\enc_e(E_A)) - E_A\| \\
 \label{equ:lin_latent_space2}
  L_{lin}(E_A, E_B) &=\left\| \dfrac{\dec_e(\enc_e(E_A)) + \dec_e(\enc_e(E_B))}{2} 
 - \dec_e \left( \dfrac{\enc_e(E_A) + \enc_e(E_B)}{2} \right) \right\|^2.
\end{align}

Here $E_A, E_B$ are the lists of edge lengths corresponding to the triangle meshes $A, B$ given during training. Our motivation for the second loss $L_{lin}$ is to explicitly encourage linear structure, which promotes smoothness of interpolated edge lengths and thus, as we demonstrate bellow, minimizes intrinsic distortion.

\vspace{-2mm}
\subsubsection{Mapping networks} Given two pretrained auto-encoders described above, we train two dense mapping networks that translate elements between the two latent spaces. We use $M_{PE}$ and $M_{EP}$ to denote the networks that translate an element from the shape (resp. edge) latent space to the edge (resp. shape) latent space.

To define the losses we use to train these two networks, for a training mesh $A$ we let $l_A = \enc_p(A)$ denote the latent vector associated with $A$ by the shape encoder. Recall that when training the shape AE we compare $A$ with $\dec_p(l_A)$. To train our mapping networks $M_{PE}$ and $M_{EP}$ we instead compare $A$ with $ \dec_{p}\left(M_{EP}(M_{PE}(l_A)\right)$. In other words, rather than decoding directly from $l_A$ we first map it to the edge length latent space (via $M_{PE}$). We then map the result back to the shape latent space (via $M_{EP}$) and finally decode the 3D shape. We denote the shape reconstructed this way by $\tilde{A} = \dec_{p}(M_{EP}(M_{PE}(\enc_{p}(A)))).$ We compare $\tilde{A}$ to the original shape $A$, which leads to the following loss:
\begin{align}\label{equ:lmap1}
L_{map1}(A) &= d^{\text{rot}}(\tilde{A}, A).
\end{align}
Here $d^{\text{rot}}$ is a \emph{rotation invariant} shape distance comparing the the original and reconstructed shape. We use it since the list of edge lengths can only encode a shape up to rigid motion \cite{gluck1975almost}. Specifically, we first compute the optimal rigid transformation between the input shape $A$ and the predicted point cloud $\tilde{A}$ using Kabsh algorithm~\cite{arun1987least}. We then compute the mean square error between the coordinates after alignment.
 As shown in \cite{huang2019operatornet} this loss is differentiable using the derivative of the Singular Value Decomposition.

 Our second loss compares the edge lengths of the reconstructed shape $\tilde{A}$ to the edge lengths of $A$. For this we use the standard $L_2$ norm:
\begin{align}\label{equ:lmap2}
L_{map2}(A) &= \|E_{A} - E_{\tilde{A}}\|_2^2,
\end{align}
where $E_{A}$ denotes the list of edge lengths of shape $A$.

Our last loss considers a similar difference but starting in the edge length latent space, rather than the shape one. Specifically, given a shape $A$ with list of edge lengths $E_A$, we first encode it to the edge length latent space via $\enc_{e}(E_A)$. We then translate the resulting latent vector to the shape latent space (via $M_{EP}$) and back to the edge length latent space (via $M_{PE}$), and finally decode the result using $\dec_e$. This leads to the following loss:
\begin{align}\label{equ:lmap3}
L_{map3}(A) &= \| \dec_{e}(M_{PE}(M_{EP}(\enc_{e}(E_A)))) - E_A \|^2_2,
\end{align}
 
Our overall loss  is then simply a weighted sum of three terms $\alpha L_{map1} + \beta L_{map2} + \gamma L_{map3}$ for shapes given at training where $\gamma$ is non-zero.

\subsubsection{Network Training} 
To summarize, we train our overall network architecture described in Figure \ref{fig:architecture} in three separate steps. First we train the shape-based auto-encoder using the loss given in Eq. \eqref{equ:shape_ae_loss}. Then we train the edge length auto-encoder using the sum of the losses in Eq. \eqref{equ:lin_latent_space1} and Eq. \eqref{equ:lin_latent_space2}. Finally we train the dense networks $M_{EP}$ and $M_{PE}$ using the sum of the three losses in Eq. \eqref{equ:lmap1}, Eq. \eqref{equ:lmap2}, Eq. \eqref{equ:lmap3}. We also experimented with training the different components jointly but have observed that the problem is both more difficult and the relative properties of the computed latent spaces become less pronounced when trained together, leading to less realistic reconstructions.

\subsection{Navigating the restricted latent space}
After training the networks as described above, we use them at test time for shape reconstruction and interpolation. We stress that at test time we do not use the edge encoder and decoder networks $ \enc_e $, $ \dec_e $, as they require canonical edge ordering. Instead we use the permutation invariant shape based auto-encoder and the mapping networks $M_{PE}, M_{EP}$ to better preserve intrinsic shape properties.

Our main observation is that the latent space associated with the shape auto-encoder provides a way to recover realistic point clouds, while the latent space of the edge length auto-encoder helps to impose a better distance structure in that space. Note that our approach is related to methods for reconstructing a shape from its edge lengths, which while possible theoretically \cite{gluck1975almost}, is computationally challenging and error prone in practice \cite{wang2012linear,boscaini2015shape,corman2017functional,chern2018shape}. By using a learned shape space, however, our reconstruction is both efficient and leads to realistic shapes.

\subsubsection{Interpolation} Given two possibly noisy unorganized point clouds $P_A$ and $P_B$ we first compute their associated edge-based latent codes: $m_A = M_{PE}(\enc_{p}(P_A))$ and $m_B = M_{PE}(\enc_{p}(P_B))$. Here we use the permutation-invariance of our encoder $\enc_{p}$ allowing to encode unordered point sets. We then linearly interpolate between $m_A$ and $m_B$ but use the \emph{shape decoder} $\dec_p$ for reconstruction. Thus, we compute a family of intermediate point clouds as follows:
\begin{equation}\label{equ:interpolation}
{P_{\alpha}} = \dec_{p}\left( M_{EP}\left( (1-\alpha) m_A + \alpha m_B\right) \right), ~ \alpha \in [0\ldots 1]
\end{equation}
In other words, we interpolate the latent codes in the edge-based latent space, but perform the reconstruction via the shape decoder $\dec_p$. This allows us to make sure that the reconstructed shapes are both realistic and their intrinsic metric is interpolated smoothly. Note that unlike the purely geometric methods, such as \cite{kilian2007geometric}, our approach does not rely on the given mesh structure at test time. Instead, we employ the learned edge-based latent space as a proxy for recovering the intrinsic shape structure, which as we show bellow, is sufficient to obtain accurate and smooth interpolations.

Since the edge length auto-encoder is fully rotation invariant, it is necessary to align the output shapes at test time. We can do so easily by using the same optimal rigid transformation as used to compute Eq. \eqref{equ:lmap3}.  
\subsection{Unsupervised training}

Our method can be adapted to an unsupervised context where the 1-1 correspondences are not provided during training. Contrary to our main pipeline, we cannot compute the edge lengths directly from the training data. However, we can encourage the model to produce a consistent mesh as described in \cite{groueix20183d}. We initialize the weights by pre-training on a selected mesh using the reconstruction loss $L_{rec}$ described in \eqref{equ:shape_ae_loss} and train the model using Chamfer distance and regularization losses to keep the triangulation consistent. Finally, we can train the edge-length auto-encoder by using the output of the shape auto-encoder as training data. We describe this process in detail in \ref{sec_supp:unsupervised}.       







\section{Results}
\label{sec:results}

\subsubsection{Datasets}
We train our networks on two different datasets: humans and animals. For  humans,  we use the dataset proposed in \cite{huang2019operatornet}. The dataset contains 17440 shapes subsampled to 1k points from DFAUST \cite{dfaust:CVPR:2017} and SURREAL \cite{varol17_surreal}. The test set contains 10 sub-collections (character
+ action sequence, each consisting of 80 shapes) that are
isolated from the training set of DFAUST and 2000 shapes from SURREAL dataset.
During training the area of each shape is normalized to a common value. For animals we sample 12000 shapes from the SMAL dataset \cite{Zuffi:CVPR:2017}. We sample an equal number of shapes from the 5 categories (big cats, horses, cows, hippos, dogs) to build a training set of 10000 shapes and a testset of 2000 shapes. We simplify the shapes from SMAL to 2002 points per mesh. The animal dataset provides challenging shape pairs that are far from being isometric, some of which we highlight in provided video.

\subsection{Shape interpolation}
We evaluate our method on our core application of shape interpolation and compare against six different recent baselines. Namely, we compare to 
three data-driven methods, by performing linear interpolations in the latent spaces of auto-encoders using PointNet \cite{qi2017pointnet} and PointNet++ \cite{qi2017pointnet++} architectures as well as the pre-trained auto-encoder proposed in the state-of-the-art non-rigid shape matching method 3D-CODED \cite{groueix20183d}. 

We also compare to three optimization-based geometric methods, by building on the ideas from \cite{kilian2007geometric,shao2018riemannian,chen2017metrics}.
We produce our first two baselines by initializing a linear path in latent space of our shape auto-encoder and optimizing each sample via 1000 steps of gradient descent. We use GD EL to denote the method that optimizes $E_{disc}$ as described in Eq.\eqref{eq:disc_opt}, and G2 L2 to denote the method that minimizes the L2  variance over the interpolated shape coordinates as described in \cite{shao2018riemannian}. 
Finally  we compare to a method simplified from \cite{kilian2007geometric} (GD Coord.), in which we first initialize a path by linearly interpolating the coordinates of  source and target shapes. Similarly to GD EL, we minimize the discrete interpolation energy $E_{disc}$ using gradient descent on the point coordinates directly. 

Remark that GD Coord., GD L2 and GD EL methods all rely on gradient descent to compute each interpolation \emph{at test time}. In other words, these approaches all require to solve a highly non-trivial optimization problem during interpolation, leading to additional computational cost and parameters (learning rate, number of iterations). In contrast our method outputs a smooth interpolation in a single pass.

\begin{table}[]
\begin{center}
 \begin{tabular}{l||l|l|l|l||l|l|l}
 & \multicolumn{4}{c||}{Direct inference} & \multicolumn{3}{c}{Optimization based} \\ \hline
                 & Ours & PointNet & 3D-Coded & PointNet++ & GD L2 & GD EL & GD Coord. \\ \hline
EL                &\textbf{0.2311}&0.3510 &  0.6130 & 0.2993 & 0.3631 & 0.2985& \textbf{0.0345}\\
Area ($10^{-4}$)   &\textbf{1.261}&1.773& 3.137  & 1.586 & 1.838 &1.714 & \textbf{0.248}\\
Volume ($10^{-4}$) & \textbf{0.342}& 1.613& 1.243 & 335.2 & 1.483 & 1.703&\textbf{0.152}
\end{tabular}
\caption{We report the mean squared variance of the edge length (EL), per surface area and total shape volume over the interpolations of 100 shape pairs. We highlight that among the direct inference methods our method achieves lowest variance across all intrinsic features. We highlight the best numerical results per category. GD coord. leads to interpolation with low distortion, as it optimizes the coordinates directly however the shapes are not realistic (see Figure \ref{fig:interpolation_comp})}
\label{tab:interpolation}
\end{center}
\end{table}


To evaluate the interpolations we sample 50 shapes from the DFAUST testset using farthest point sampling. We then test on 100 random pairs from those 50 shapes. We use our pipeline trained with $\alpha = 30$, $\beta = 1200$ and $\gamma=800$ in the mapping networks loss described in \ref{sec:architecture}. We provide an ablation study on the choice of losses in \ref{sec_supp:choice_loss}.

 Table~\ref{tab:interpolation} shows quantitative comparisons. Given an interpolation path $(S_n)$ obtained by each method, we compute the mean squared variance of various shape features $f$ on the path. We consider three features: lengths of edges, overall surface area and overall volume  enclosed by the shape (computed from the mesh embedding). For each of these, we compute the sum of the squared differences across all instances in the interpolating sequence:

\begin{equation}\label{eq:quant_interpolation}
 Var_f(S_n) = \frac{1}{n-1}\sum_{i=2}^{n} \|f(S_{i}) - f(S_{i-1})\|^2. 
\end{equation}

Intuitively, we expect a good interpolation method to result in smooth interpolations which would have low variance across all of the intrinsic shape properties. To be fair when comparing with PointNet++ since it was trained on normalized bounding boxes and not area, we normalize the total area of each output. The large volume variance of this baseline is primarily due to bad reconstruction quality of the source and target shapes.

\begin{figure*}[t]
\begin{center}

   \includegraphics[width=1\linewidth]{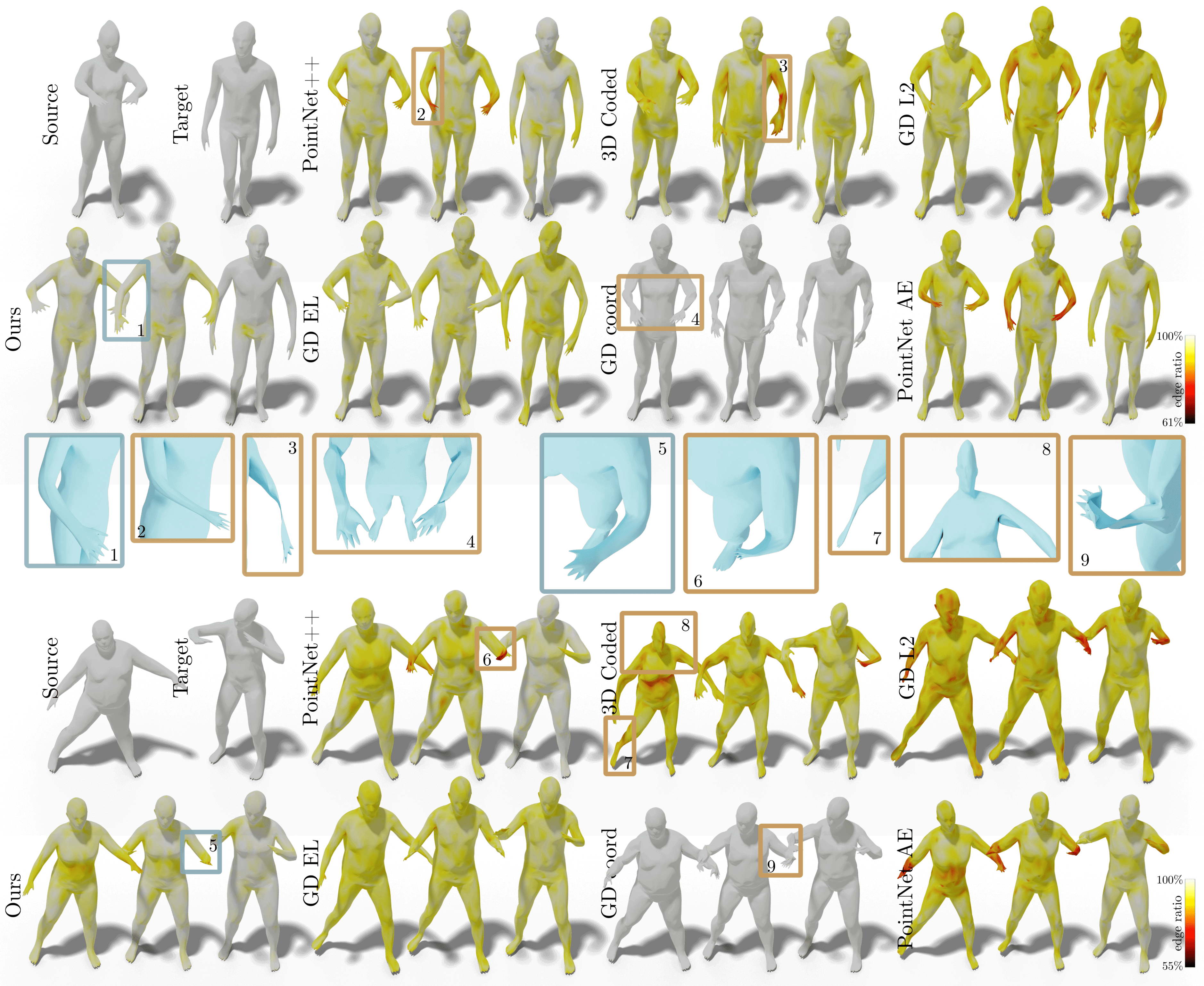}
\end{center}
   \caption{Qualitative comparison of interpolation on DFAUST testset. We display the edge ratio between the linear interpolation of the target and source edges and the produced interpolation.}
\label{fig:interpolation_comp}

\end{figure*}

As shown in Table~\ref{tab:interpolation} our method produces the best results among the direct data-driven methods and  the best results over all the baselines except from GD Coord. This latter method is not data-driven and optimizes edge lengths directly on the coordinates without any constraints. As such it produces shapes with low distortion but that are not realistic (see Figure \ref{fig:interpolation_comp}). Furthermore, similarly to \cite{kilian2007geometric} it requires the input shapes to be represented as meshes in 1-1 correspondence.


In all qualitative figures, we visualize the minimum ratio between the linear interpolation of the ground truth edge lengths and the edge lengths of the produced shapes. We color-code this ratio to highlight areas of highest intrinsic distortion (shown in red).



\begin{figure}
\begin{center}

   \includegraphics[width=1\linewidth]{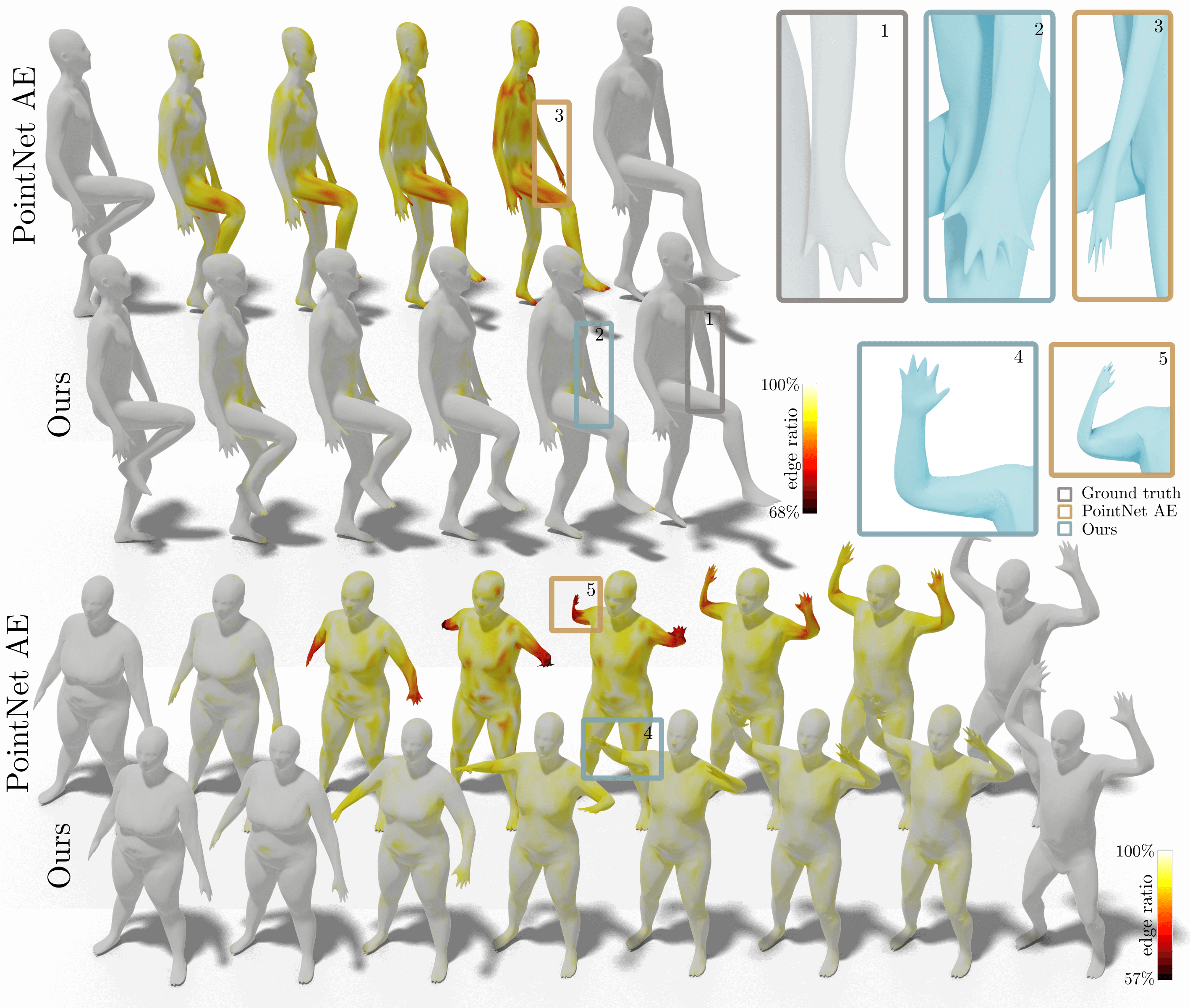}
\end{center}
\caption{We compare linear interpolations in PointNet AE latent space
  and interpolation using our approach. We visualize the ratio
  between the linear interpolation of edge lengths and edge lengths of
  the computed interpolations, to help highlight problematic areas.}
\label{fig:interpolation}
\end{figure}

In Figure~\ref{fig:interpolation} we provide qualitative
comparison of the linear interpolations in the basic shape (PointNet)
AE latent space and the interpolation using our method. Our method
preserves body type better (row 2) and interpolates well between a
pair of shapes where the end results differs highly from the linear
interpolation of the coordinates (row 4).

In Figure \ref{fig:interpolation_comp} we illustrate the interpolated shapes between the input source and target, shown in grey. We observe that PointNet AE and PointNet++ methods tend to produce results that are closer to linear interpolation of the coordinates. As highlighted above, we notice that while GD Coord. has low variance in the interpolated intrinsic features, the reconstructed shapes do not look natural. Overall, our method presents less distortions and more smooth interpolations compared to all baselines. We present more comparisons and evaluations in the provided video.

\begin{figure*}[t]
\begin{center}

   \includegraphics[width=0.9\linewidth]{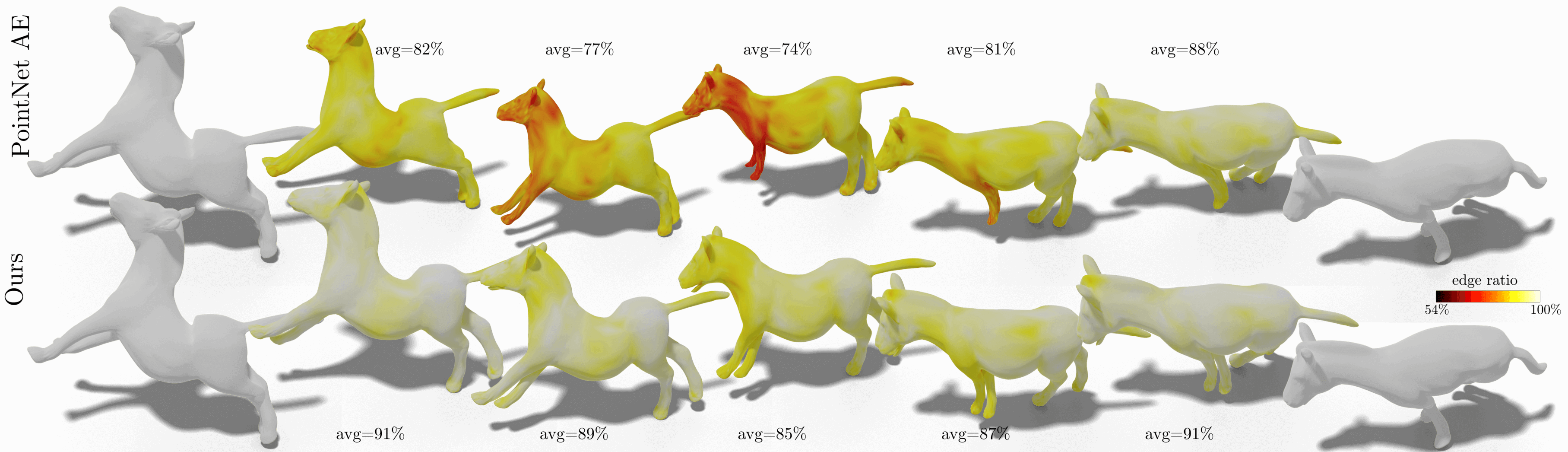}
\end{center}
   \caption{Interpolation of two horses from SMAL dataset}
\label{fig:interpolation_smal}

\end{figure*}

We further evaluate our model on the SMAL dataset. To build the interpolation pairs from the test set, we sample 10 shapes per category by farthest points sampling. We then choose 100 random pairs from that dataset.
In Figure~\ref{fig:interpolation_smal} we show results of interpolating between two horses. We observe that linear interpolation in the shape latent space leads to shape distortions such as shorter legs (middle) and wrong shape size estimation (top left). 
The Shape AE (resp. Ours) produces a edge variance of 2.068 (resp. 1.548). Similarly to above, our method shows improvement at interpolating intrinsic information. We provide detailed numerical evaluation of interpolations on SMAL in \ref{sec_supp:additional}.

\subsubsection{Interpolation in the unsupervised case.} The unsupervised Shape AE (resp. Ours) produces a edge variance of 0.599 (resp. 0.394). While we observe better results in the supervised setting, our method nevertheless produces quantitative and qualitative improvement over the linear interpolation in latent space. We provide further numerical and qualitative results in \ref{sec_supp:unsupervised}.

\subsection{Shape reconstruction}
\label{sec:results_reconstruction}


\begin{table}
\centering
\makebox[0pt][c]{\parbox{1.0\textwidth}{%
    \begin{minipage}[b]{0.48\hsize}\centering
        \begin{tabular}{l|c|c|c}
                 &  EL & PC   & area \\
                  &  ($10^{-5}$) &  ($10^{-4}$)  &  ($10^{-8}$)\\\hline
          PointNet AE  & 3.023 & \textbf{2.120} &2.454\\
          Edge Length AE      & 3.127 & -& - \\
        Ours  &1.641 & 2.572 & 1.562
        \end{tabular}
        \caption{Mean squared reconstruction losses on the  humans testset. Edge length reconstruction loss (EL), Point cloud coordinates reconstruction loss (PC) and per triangle area difference}
        \label{tab:recons}
    \end{minipage}
    \hfill
    \begin{minipage}[b]{0.48\hsize}\centering
      \begin{tabular}{l|c|c|c}
                 &  CD &volume & area\\
                 &  ($10^{-3}$)&  ($10^{-5}$)&\\ \hline
Shape AE & 4.703 &30.851&0.1382 \\
Ours & 4.135&9.47&0.047
\end{tabular}
        \caption{Reconstruction accuracy on SCAPE dataset. We measure the Chamfer distance (CD), mean square total volume difference and MS total area difference}
        \label{tab:scape_recons}
    \end{minipage}%
}}
\end{table}


For our method, given an unordered point cloud $P$,
we reconstruct the shapes by using the following combination of our
trained networks $\dec_{p} (M_{EP} (M_{PE} (\enc_p (P))))$, which
differs from the standard auto-encoder approach $\dec_{p}
(\enc_{p}(P))$. Therefore, in this section we show that the additional
regularization provided by our mapping networks $M_{EP}, M_{PE}$
results in better shape reconstruction.

We evaluate the reconstruction accuracy of our model on the DFAUST/SURREAL testset.  In  Table~\ref{tab:recons}, we compare the reconstruction accuracy to the base models. We measure intrinsic features: edge length and area per triangle $L2$ reconstruction loss, and extrinsic features the $L2$ coordinates reconstruction loss. Our method reconstructs the input shape intrinsic features better that the PointNet AE while producing comparable extrinsic reconstruction loss. 
 

We further evaluate the generalization capacity of our network by evaluating on the SCAPE \cite{anguelov2005scape} dataset. For testing we sample 1000 random points from the surface of each mesh. Table~\ref{tab:scape_recons} shows an improvement in the reconstruction for our method. We observe even higher relative performance when comparing the total volume and total area of the reconstructed shapes which give a sense of the perceived quality of the shapes. Shape distortions are often related to shrunk or disproportional body parts. 

\begin{figure*}[t]
\begin{center}
   \includegraphics[width=1\linewidth]{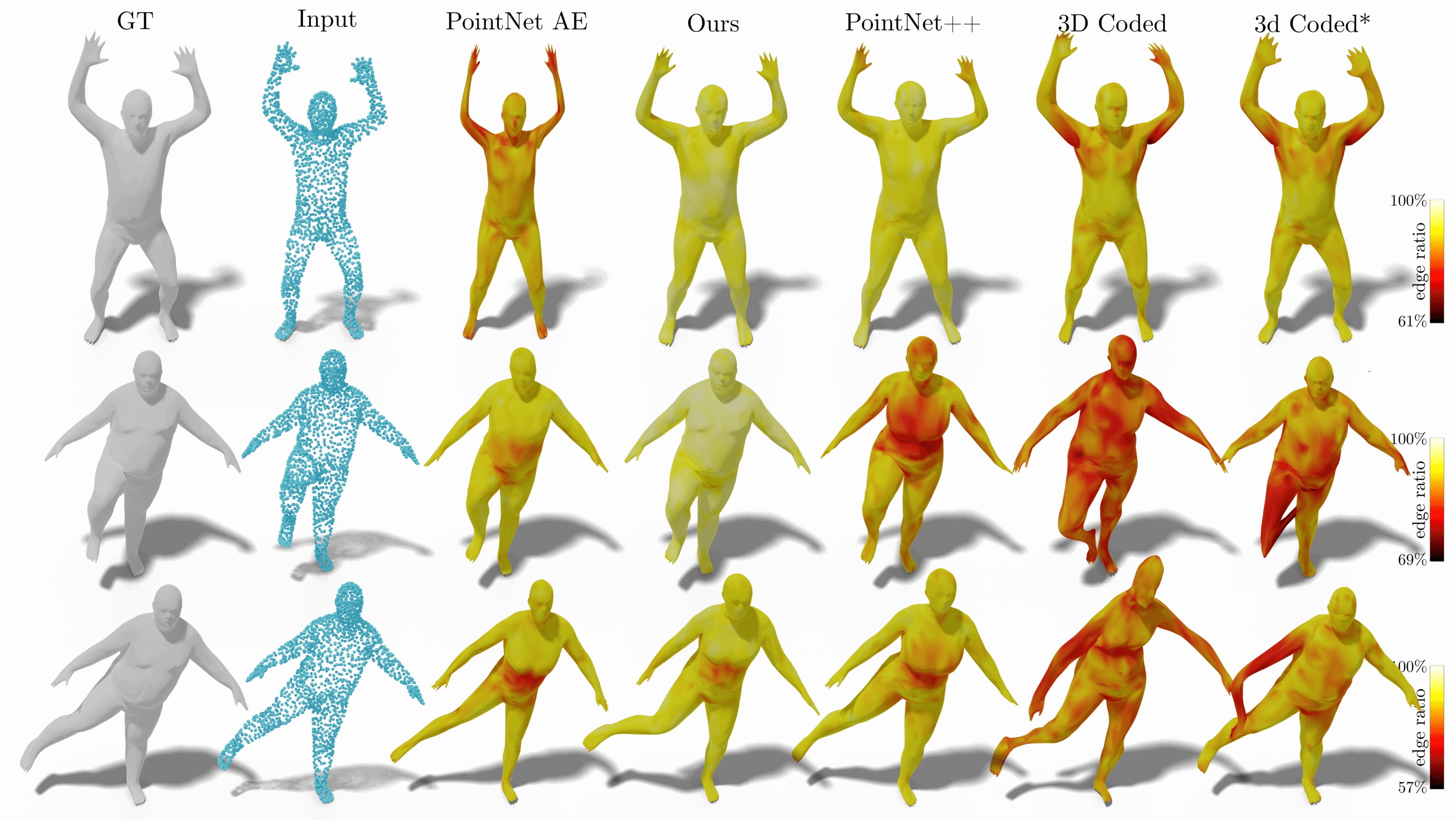}
\end{center}
   \caption{Reconstruction of meshes from point clouds containing 1000
     points, sampled from the underlying shape.}
\label{fig:reconstruction}
\end{figure*}

We show qualitative results on reconstruction in \ref{fig:reconstruction} on meshes from the DFAUST testset. To be fair to 3D-CODED, we normalize the total area of the output
shapes.  We evaluate this method before (3D-CODED) and after
(3D-CODED*) their additional step of Chamfer Distance
minimization. Note that in the case of 3D-CODED* additional
optimization \emph{at test time} is required to recompute the latent
code that best approximates the input. Our method, on the other hand,
performs the reconstruction in one shot. Overall, our method produces more precise and natural reconstructions.

Finally, as shown in Figure~\ref{fig:teaser}, our method is robust to high levels of noise (left), holes, and missing parts (right). We provide further reconstruction examples \ref{sec_supp:reconst}.

\section{Conclusion, Limitations \& Future Work}
We presented a method for interpolating unorganized point clouds. Key
to our approach is a dual latent space encoding that both captures the
overall shape structure and the intrinsic shape information, given by
edge lengths provided during training. We demonstrate that our
approach leads to significant improvement compared to existing
methods, both in terms of interpolation smoothness and quality of the
generated results. In the future, we plan to extend our method to also
incorporate other features such as semantic classes or
segmentations. It would also be interesting to explore the utility of
our dual encoding space in other applications, on images or graphs.

%
%
\bibliographystyle{splncs04}
\bibliography{egbib}

\appendix
\section*{Supplementary}

\section{Overview}
In Section \ref{sec_supp:interp} we provide additional illustrations of our
shape interpolation method. In Section \ref{sec_supp:reconst} we
demonstrate the performance of our approach for \emph{shape
  reconstruction} highlighting the utility our dual network for strong
regularization of recovering high-quality shapes from noisy point
clouds, as mentioned in the main manuscript. In Section
\ref{sec_supp:ablation} we provide an in-depth ablation study of our
network design. In Section \ref{sec_supp:unsupervised} we demonstrate the
performance of our approach in the unsupervised case (when the
training data is not in correspondence). Finally, in Section
\ref{sec_supp:architecture} we provide details of our architecture.

\section{Shape interpolation}
\label{sec_supp:interp}

\subsection{Video and Comparison to Optimization-based Approaches}

We provide a video which contains qualitative comparisons of interpolations on DFAUST and SMAL test sets with our main baselines. Note that our approach produces visually smoother interpolations with significantly lower distortions than all baselines across all shape pairs. 

In the video we also provide comparisons with optimization-based approaches that achieve low distortion in Table \ref{tab:interpolation} of the main manuscript. Specifically note that methods such as GD Coord. 1) require the input    shapes to be in 1-1 correspondence 2) rely on expensive optimization at test time (for this reason, we compute these interpolations at half of the frame-rate), and most importantly 3), as shown in the accompanying video, as they are not learning-based, lead to non-realistic intermediate shapes. 

\subsection{Additional  Evaluation}
\label{sec_supp:additional}

We further compare our method to the PointNet AE on the SMAL animals
dataset. Table \ref{tab:interpolation_comp_smal} reports the
mean-squared variance of several shape features during interpolation
of 100 pairs among 50 shapes obtained by farthest points sampling on
this dataset. Note that our method produces significantly better
quantitative results across all shape features

\begin{table}
\begin{center}
\begin{tabular}{l|c|c|c}
                 &  edge length &  area ($10^{-3}$) &volume ($10^{-2}$)  \\\hline
PointNet  &2.068& 3.742  & 2.754\\
Ours  & \textbf{1.538} & \textbf{2.975}&\textbf{1.728}
\end{tabular}
\caption{MS variance of various shape features obtained from interpolating 100 pairs among 50 shapes obtained by farthest points sampling on animals dataset (SMAL) }
\label{tab:interpolation_comp_smal}
\end{center}
\end{table}
 
\section{Shape reconstruction}
\label{sec_supp:reconst}

As mentioned in the main manuscript, our approach not only enables
better interpolation, but also results in more accurate
reconstructions from noisy input. Here we provide additional
qualitative and quantitative evaluation of the reconstruction
performance and comparison to different baseline methods.



In all of the experiments the training data is the combination of
DFAUST and SURREAL datasets, and the test data is the DFAUST test
shapes, both with and without noise.

Table~\ref{tab:recons} shows reconstruction results for several
baselines on the 800 DFAUST test shapes. We report the edge length
accuracy (EL), rotation-invariant point cloud reconstruction accuracy
(PC) and per triangle area reconstruction accuracy (area). Note that
our approach achieves the best overall reconstruction accuracy,
especially on the intrinsic quantities and gives slightly worse
reconstruction extrinsic loss (PC) compared to PointNet AE.   We
provide qualitative examples in Figure \ref{fig:reconstruction}. Note
that our method leads to both preservation of the overall shape
structure and significantly less intrinsic distortion compared to all
baselines.


\begin{table}[]
\begin{center}
\begin{tabular}{l|c|c|c}
                 &  EL ($10^{-5}$) & PC  ($10^{-4}$)  & area ($10^{-8}$) \\\hline
PointNet AE  & 3.023 & \textbf{2.120} &2.454\\
Edge Length AE      & 3.127 & -& - \\
Ours  $L_{1,2,3}$ &\textbf{1.641} & 2.572 & \textbf{1.562} \\
3D-CODED &6.323 &5.803 &5.485 \\
3D-CODED* &6.284 &4.260 &5.409 \\
PointNet++ &2.835 &3.224 &2.835\\
\end{tabular}
\caption{Mean squared reconstruction losses on DFAUST testset. Edge length reconstruction loss (EL), Point cloud coordinates reconstruction loss (PC) and per triangle area difference}
\label{tab:recons}
\end{center}
\end{table}



\begin{figure*}[t]
\begin{center}

   \includegraphics[width=1\linewidth]{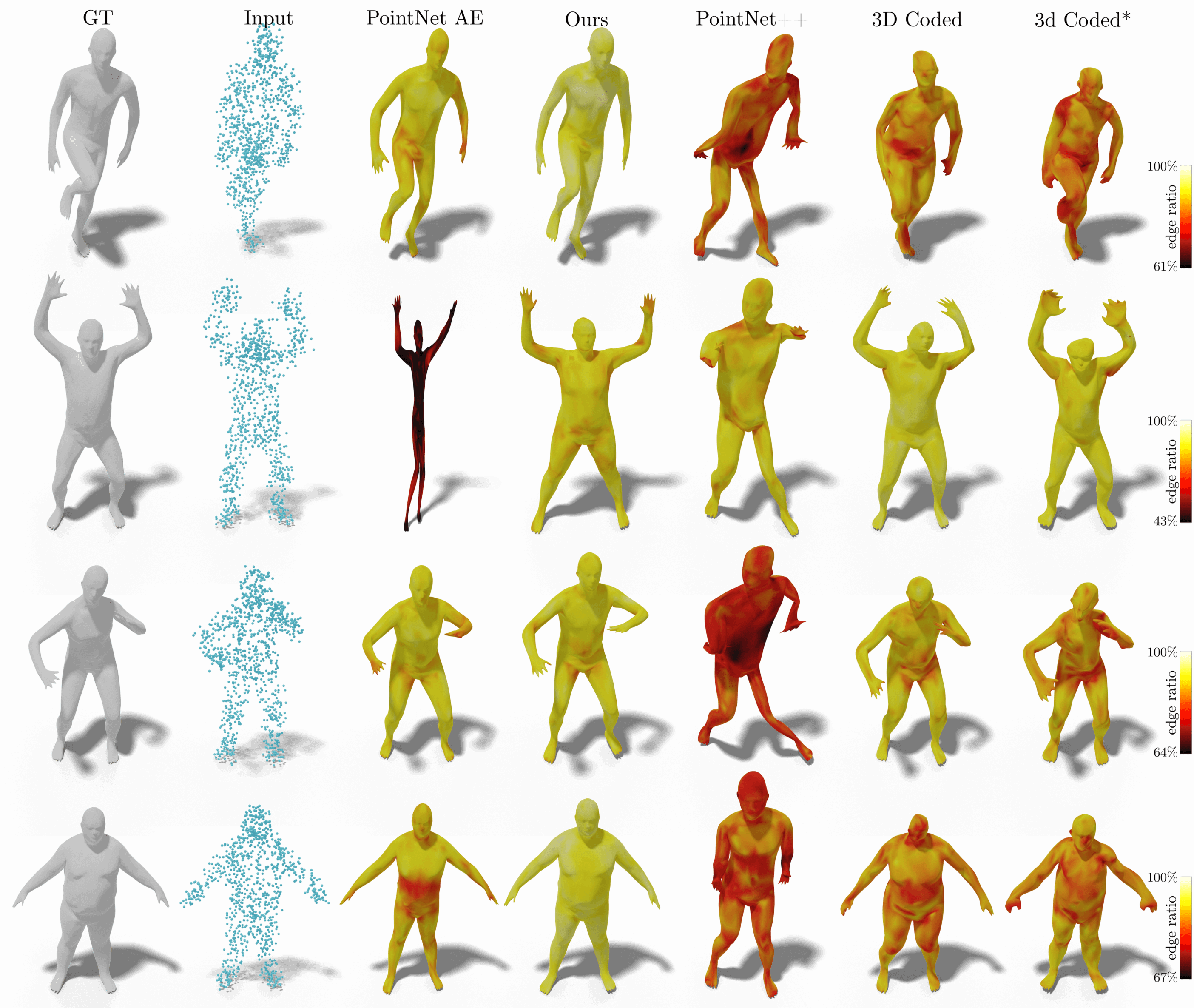}
\end{center}
   \caption{Reconstructions from point clouds with 5\% of the shape scale gaussian noise. }
\label{fig:reconstruction_noise}

\end{figure*}

\begin{figure*}[t]
\begin{center}

   \includegraphics[width=0.9\linewidth]{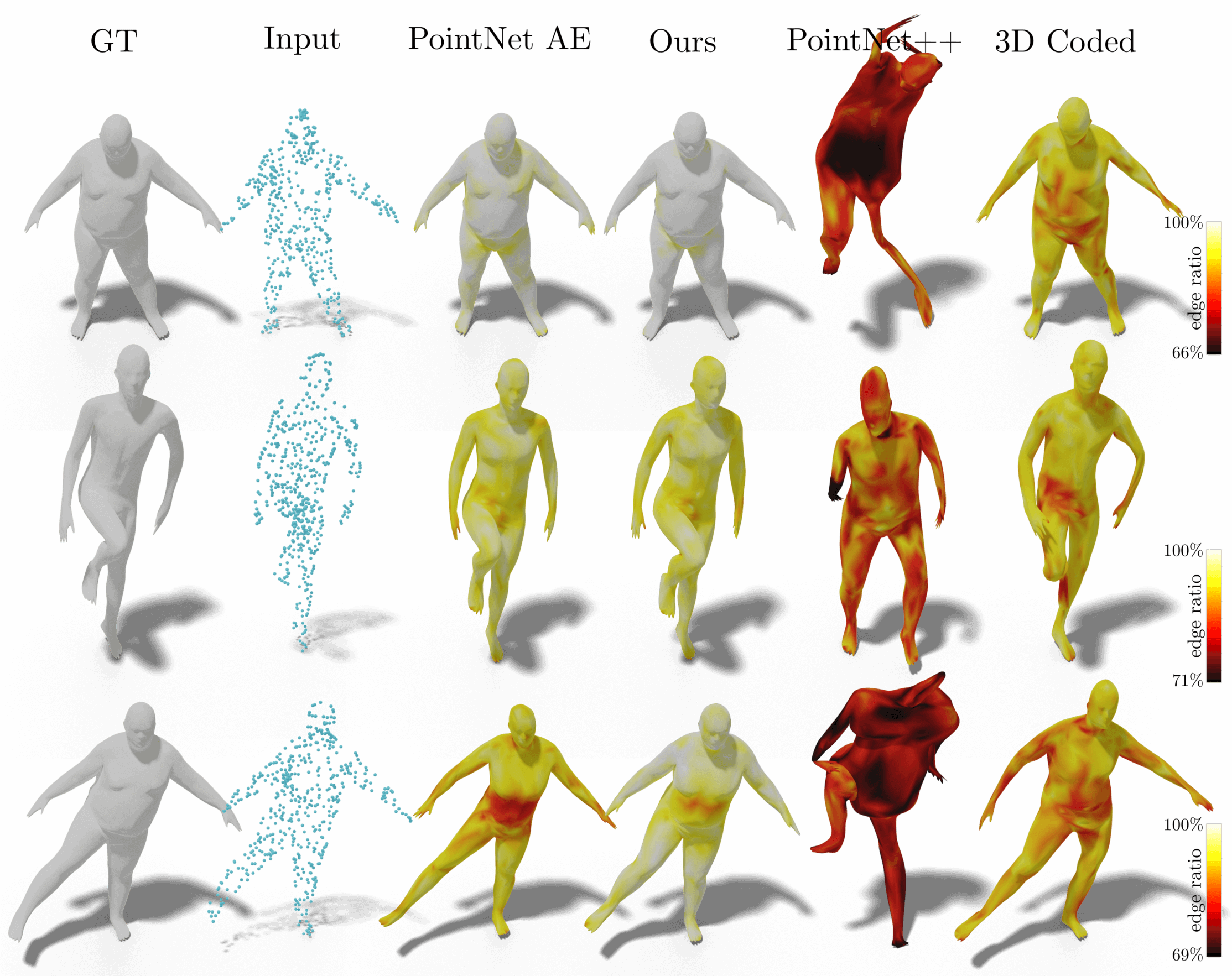}
\end{center}
   \caption{We reconstruct a mesh from 500 points sub-sampled randomly from the ground truth mesh. We use a network pre-trained on inputs of size 1000 points. }
\label{fig:reconstruction500}

\end{figure*}

Table~\ref{tab:reconsnoise_sample} (left) shows reconstruction
performance on noisy point clouds. Note that we test using our model which was trained on clean data. Each noisy point cloud is obtained by adding Gaussian noise
magnitude 5\% of the scale of the mesh to each vertex coordinate. We
observe that our method outperforms the other baselines for all the
features. Figure \ref{fig:reconstruction_noise} shows reconstructed
meshes from the noisy point clouds. Figure \ref{fig:reconstruction500} shows reconstructed meshes from point clouds under-sampled to only 500 points.  Notice that our method performs
better at recovering the original pose and body type than the
different baselines.

\begin{table}[]
\centering
\begin{tabular}{l||c|c|c||c|c|c}
& \multicolumn{3}{c||}{Noisy dataset} & \multicolumn{3}{c}{Undersampled dataset} \\ \hline
                 &  EL ($10^{-5}$) & PC  ($10^{-4}$)  & area ($10^{-8}$) &  EL ($10^{-5}$) & PC  ($10^{-4}$)  & area ($10^{-8}$) \\\hline
PointNet AE  &5.663&  8.538 &5.650  &3.847 &  \textbf{3.313} &2.810\\
Ours &\textbf{3.016} & \textbf{7.329}&  \textbf{2.812} &\textbf{1.854} & 3.587 &  \textbf{1.685}\\
3D-CODED & 8.553&10.463& 7.058 &6.219&6.898 &5.341\\
PointNet++ &26.837 &81.379&18.23 & 36.223&117.824 &27.541\\
\end{tabular}
\caption{Mean squared reconstruction losses on the DFAUST testset with noise (left) or undersampled (right). We use $5\%$ of the shape bounding box gaussian noise on the testset. We randomly sample 500 points from the test shapes surfaces. We recall that the network was trained on 1000 point clouds. We show the edge length reconstruction loss (EL), the rotation invariant reconstruction loss (PC) and the per triangle area difference}
\label{tab:reconsnoise_sample}
\end{table}


Table~\ref{tab:reconsnoise_sample} (right) shows reconstruction results on simplified point clouds. We randomly sample 500 points from the test shapes surfaces. We recall that the network was trained on 1000 point clouds. We observe that our method is more robust to under-sampling. In particular, and contrary to other methods, the intrinsic properties remain competitive with the performance from Table \ref{tab:recons}.



\begin{figure}[t]
\begin{center}

  \includegraphics[width=1\linewidth]{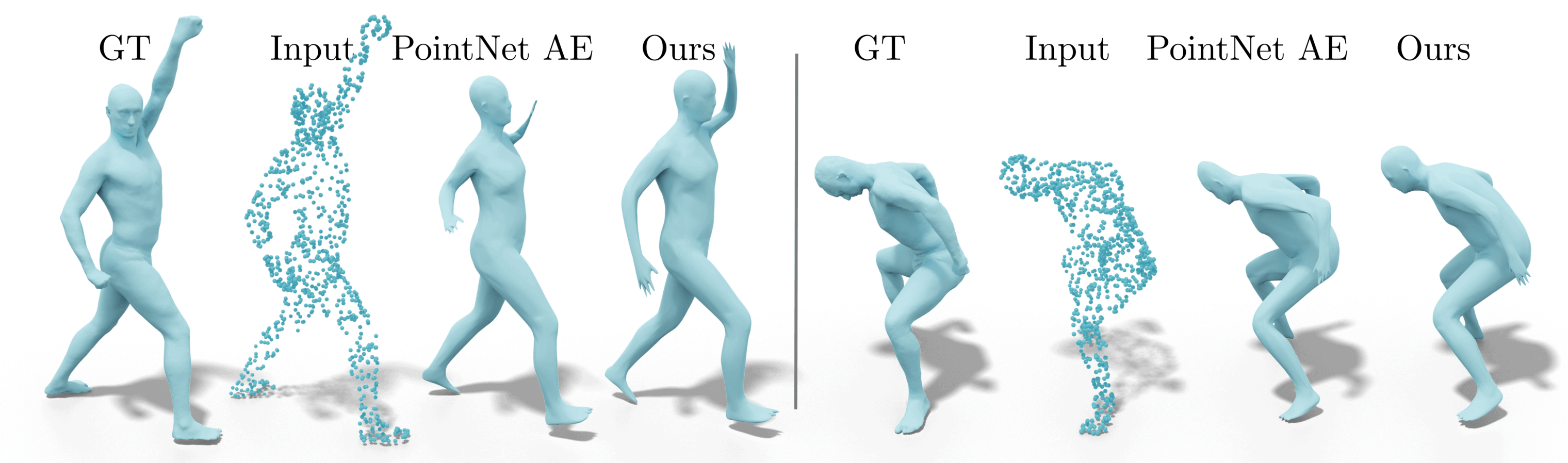}
\end{center}
  \caption{Shape reconstruction from SCAPE. We reconstruct from 1k random points on the surface.}
\label{fig:recons_scape}

\end{figure}

We also demonstrate the generalization power across different datasets
by showing in Figure~\ref{fig:recons_scape} examples of
reconstructions from SCAPE dataset \cite{anguelov2005scape}. While the
simple PointNet AE, is still able to reconstruct the overall position
of the tested human, the output has distortions near the hands (left)
and the legs (right). Our method generates more natural meshes even
though the dataset is completely unknown with an entirely different
underlying mesh, different body type and poses that are different to
those seen at training. Note that we do not display the color coding
as we do not have access to ground truth edge lengths.

\section{Ablation study}
\label{sec_supp:ablation}
\subsection{Architecture design}

\subsubsection{Importance of multiple separate networks}

\begin{figure}[t]
\begin{center}
    \includegraphics[width=1\linewidth]{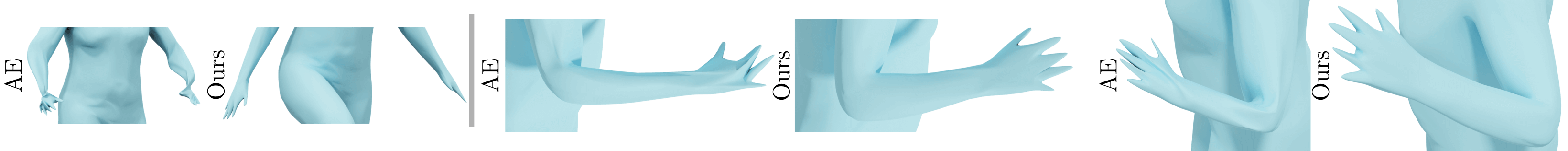}
\end{center}
   \caption{Simple AE trained with $L_e$ and $L_{rec}$ (left) or $L_e$, $L_{rec}$ and $L_{lin}$ (right)  produces artifacts during interpolation.}
\label{fig:baseline2loss}

\end{figure}

We first test the utility of having separate networks, rather than
training a single network with a combined loss. Specifically, in our study, we have
observed that introducing intrinsic information directly during the
training of the shape auto-encoder produces unrealistic results with
significant artefacts. (Fig. \ref{fig:baseline2loss}) We train two
point-cloud AE (auto-encoders) using: a combination of edge ($L_e$) and point
coordinate ($L_{rec}$) losses and edge ($L_e$), point coordinate
($L_{rec}$) and linearity losses ($L_{lin}$)

\subsubsection{Effect of separate networks training} 
\begin{figure}[t]
\begin{center}

   \includegraphics[width=0.5\linewidth]{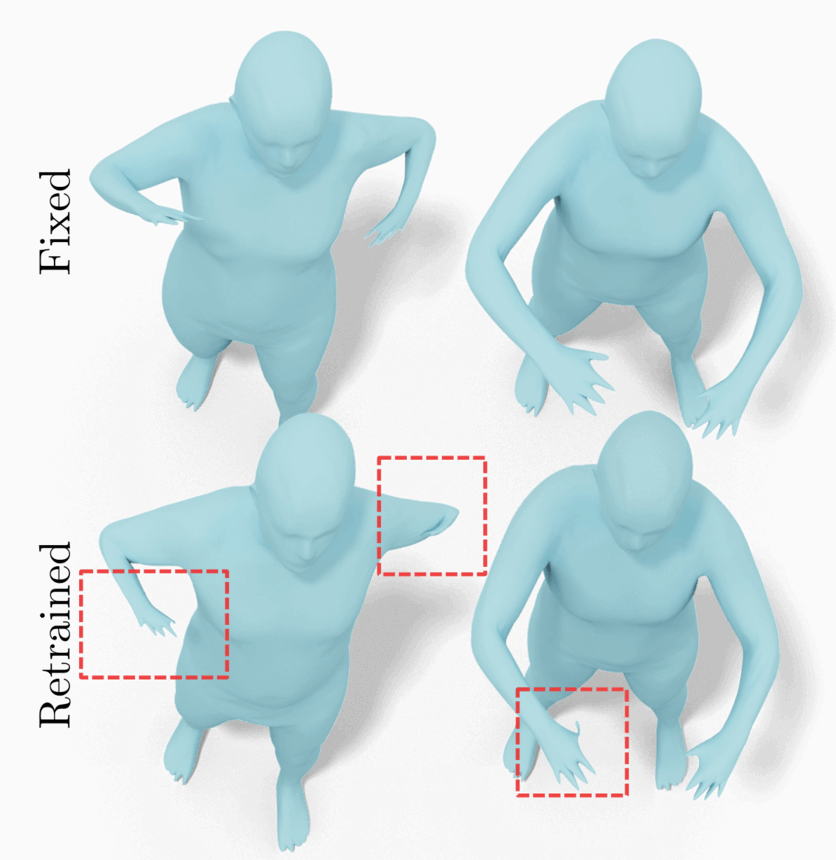}
\end{center}
   \caption{Shape distortions are appearing during interpolation if the shape AE, edge auto-encoder and mapping networks are trained at the same time.}
\label{fig:train_together}

\end{figure}

In our experiments, we fix the weights of the shape AE and edge auto-encoder during the training of the mapping networks. By doing so, we fix the latent space and generating capabilities of each network. We believe that if this constraint is not respected, the shape AE and edge auto-encoder can be indirectly trained for different losses and generate distortions in the generated shapes. Here, we train the mapping networks, edge auto-encoder and shape AE at the same time. To make the training easier, we use a pretrained shape AE and edge auto-encoder. As seen in Table~\ref{tab:recons_alltrain}, the reconstruction losses are better than before.  However, the shape AE can produce non natural reconstructions during interpolations as shown in Figure~\ref{fig:train_together}. We believe that if the shape AE and edge auto-encoder network were not pretrained, the resulting reconstructed shapes would present even more distortions since the pretrained shape AE can already generate decent natural looking shapes on parts of the dataset. 

\begin{table}[]
\centering
\begin{tabular}{l|c|c|c}
                 &  EL ($10^{-5}$) & PC  ($10^{-4}$)  & area ($10^{-8}$) \\\hline
Ours  &1.666 & 2.611 & 1.554\\
Ours sim. train. &1.027 & 1.464 & 1.027\\
\end{tabular}
\caption{Mean squared reconstruction losses on the DFAUST testset. We present our main network and an alternative model where all three components are trained simultaneously. Edge length reconstruction loss (EL), Point cloud rotation invariant reconstruction loss (PC) and per triangle area difference (area).}
\label{tab:recons_alltrain}
\end{table}

\subsubsection{Auto-encoder vs Variational auto-encoder}
During our study we compared the performances of our pipeline using either a PointNet AE or a PointNet VAE. The type of network did not result in significant differences. By instance the mean squared variance of the edge length for our architecture trained with a VAE is 0.2301 and 0.2311 when trained with a AE (respectively 0.3760 and 0.3510 for the simple VAE and AE without using our pipeline).
\subsection{Choice of losses}
\label{sec_supp:choice_loss}

\subsubsection{Importance of cycle consistency loss.} We train the mapping networks with direct reconstruction losses instead of cycle consistency losses as described in section 4.2 with $L_{map1}$, $L_{map2}$, $L_{map3}$ :

\begin{align}\label{equ:direct}
L_{direct}(P, E_P) &=\alpha \| \dec_{p}(M_{EP}(\enc_{e}(E_P)))) - P \|^2  \\
\nonumber
& + \beta \| el(\dec_{p}(M_{EP}(\enc_{e}(E_P))))) - E_P \|^2 \\
\nonumber
 &+ \| \dec_{e}(M_{PE}(\enc_{p}(P))) - E_P \|^2
\end{align}

\begin{figure}[t]
\begin{center}

   \includegraphics[width=0.6\linewidth]{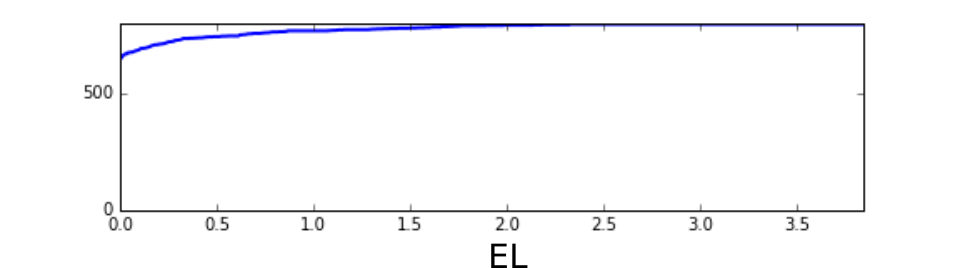}
\end{center}
   \caption{Cumulative distribution function of edge reconstruction loss on the DFAUST testset for our network trained without cycle consistency with $L_{direct}$.}
\label{fig:el_distribution}

\end{figure}

In Table~\ref{tab:reconsdirect}, we observe that the quality of the
map and the quality of the reconstructions are worse.  In
Figure~\ref{fig:el_distribution} we show the cumulative distribution
function of the edge length reconstruction loss on the testset. While
most shapes seem to have reasonable edge reconstruction quality,
outlier points make the reconstruction loss explode. Since cycle
consistency is not enforced, the network can map shapes onto outliers
in the shape space that do not correspond to reasonable natural
shapes.

\begin{table}[]
\centering
\begin{tabular}{l|c|c|c}
                 &  EL & PC  & area \\\hline
PointNet AE  & $3.023*10^{-5}$ &  $2.120*10^{-4}$ &$2.454*10^{-8}$\\
Ours  & $1.641*10^{-5}$  & $2.572*10^{-4}$ &  $1.562*10^{-8}$\\
Ours  $L_{direct}$ &0.1019 & 0.6289 &  1.338$*10^{-2}$\\
\end{tabular}
\caption{Mean squared reconstruction losses on the DFAUST testset.}
\label{tab:reconsdirect}
\end{table}

\subsubsection{Mapping losses}

In Table \ref{tab:ablation_recons} we show an ablation study of the different losses combinations (described in section 4.2 of the main manuscript) used for training the mapping networks. The subscripts $1, 2, 3$ denote the use of $L_{map1}$, $L_{map2}$, $L_{map3}$ respectively. We observe that when trained with $L_{map2}$, $L_{map3}$, so only intrinsic features, the model produces better intrinsic reconstruction performances to the expense of the extrinsic reconstruction loss. On the contrary, when trained with only $L_{map1}$ and $L_{map3}$ the network produces good point coordinate reconstruction but worse intrinsic reconstruction performances. To combine the benefits of the different losses, we choose to experiment with a model trained with the 3 losses.

\begin{table}[]
\begin{center}
\begin{tabular}{l|c|c|c}
                 &  EL ($10^{-5}$) & PC  ($10^{-4}$)  & area ($10^{-8}$) \\\hline
Ours    $L_{2,3}$         & 1.595 & 14.816& 1.490\\
Ours    $L_{1,3}$         &2.301 & 2.245& 2.113\\
Ours  $L_{1,2,3}$ &1.641 & 2.572 & 1.562 \\
\end{tabular}
\caption{Ablation study on different mapping network losses. The subscripts $1, 2, 3$ refer to $L_{map1}$, $L_{map2}$, $L_{map3}$ respectively. We show the mean squared reconstruction losses on DFAUST testset. Edge length reconstruction loss (EL), Point cloud coordinates reconstruction loss (PC) and per triangle area difference}
\label{tab:ablation_recons}
\end{center}
\end{table}

\subsubsection{Linearity regularization term in edge auto-encoder.} We train a version of our network without the linearity regularization term $L_{lin}$ described in Eq. (6) of the main manuscript for training the edge auto-encoder. As seen in  Table \ref{tab:interpolation_edge}, the interpolations  in the latent space of the edge auto-encoder are smoother when the network is trained with the linearity term. In Table \ref{tab:interpolation_ablation}, we observe that this term is also related to smoother interpolations of shapes.

\begin{table}
\centering
\makebox[0pt][c]{\parbox{1.0\textwidth}{%
    \begin{minipage}[b]{0.4\hsize}\centering
        \begin{tabular}{l|c}
                 &  EL  \\\hline
Edge AE  &0.199 \\
Edge AE no lin. reg. &1.777
\end{tabular}
\caption{We report the mean squared variance of the edge length (EL) over the interpolation in the edge length AE latent space of 100 shape pairs. }
\label{tab:interpolation_edge}

    \end{minipage}
    \hfill
    \begin{minipage}[b]{0.55\hsize}\centering
     \begin{tabular}{l|c|c|c}
                 &  EL &  area ($10^{-4}$) &volume ($10^{-4}$)  \\\hline
Ours &0.230&1.220&0.385\\
Ours no lin. reg. &0.245&1.361&0.430
\end{tabular}
\caption{Interpolation losses for our network where the edge auto-encoder is trained with and without linearity regularization term. We report the mean squared variance of the edge length (EL), per surface area and total shape volume over the interpolations of 100 shape pairs from the DFAUST testset.  }
\label{tab:interpolation_ablation}
    \end{minipage}%
}}
\end{table}





\section{Interpolation in unsupervised case}
\label{sec_supp:unsupervised}

Our method can be adapted to an unsupervised context where the 1-1 correspondences are not provided during training. The training process can be described in 3 steps: We first train a point cloud auto-encoder that takes unordered point clouds and outputs an ordered point clouds where the order corresponds to  given template $T$. Then we train the edge auto-encoder by using the output of the shape auto-encoder as training data. Finally, we train the mapping networks as described in the main manuscript.

We first initialize the weights by pre-training the shape AE network to output a chosen template mesh using a variant of the reconstruction loss $L_{rec}$ described in Eq. 4 of the main manuscript.

\begin{equation}\label{equ:shape_ae_loss}
  \vspace{-1.5mm}
  L_{recInit}(P) =\frac {1}{n}\sum_{i=1}^{n}\| T_{i}-{{\tilde{P}_{i}}}\|^{2}, \text{ where } \tilde{P} = \dec_p \left( \enc_p (P) \right).
  \vspace{-1.5mm}
 \end{equation}

Then we train the model using Chamfer Distance ($CD$) from Eq. \eqref{eq:chamfer} while encouraging the network to maintain the learned triangulation from step 1 by using regularization terms similar to those used in \cite{groueix20183d} described bellow.

\begin{equation}
    CD(\tilde{P}, P) = \frac{1}{n} \sum_{p_i \in \tilde{P}} \min_{p_j \in P} \|p_i-p_j\|^2_2 + \frac{1}{n} \sum_{p_j \in P} \min_{p_i \in \tilde{P}} \|p_j-p_i\|^2_2
\label{eq:chamfer}
\end{equation}

\begin{align}\label{equ:lin_latent_space1}
L_{e}^{reg}(E_{\tilde{P}}) &= \|E_{\tilde{P}} - E_T\|^2_2, \text{where }  \tilde{P} = \dec_p(\enc_p(P)) 
\end{align}

\begin{align}\label{equ:lin_latent_space1}
L_{lap}^{reg}(\tilde{P}) &= \| L*(\tilde{P} - T)\|^2_2 , \text{where } L \text{ is the graph laplacian}
\end{align}

\begin{figure}[t]
\begin{center}
   \includegraphics[width=0.8\linewidth]{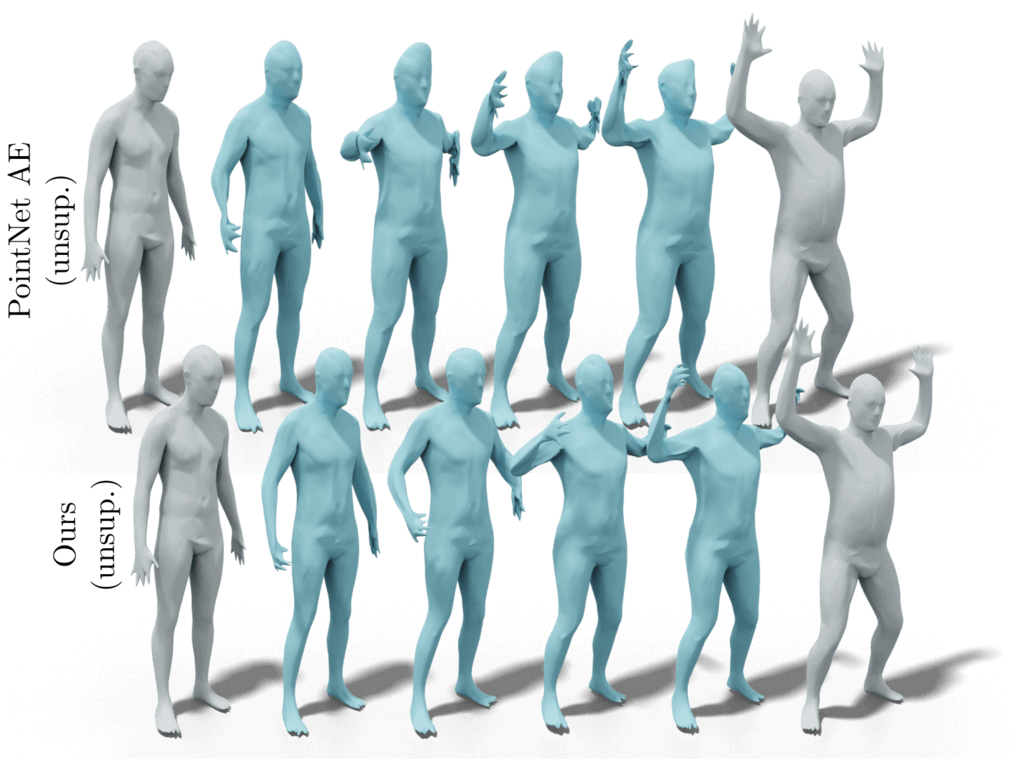}
\end{center}
   \caption{Interpolation between shapes when trained with no 1-1 correspondences at train time. Our method produces more realistic shapes.}
\label{fig:unsupervised_inter}

\end{figure}

We report numerical evaluation of the interpolations in  Table \ref{tab:interpolation_unsupervised}. Note, that our method leads to improved shape features. In Figure \ref{fig:unsupervised_inter}, we observe that our method produces more realistic shapes, in particular it produces better arms and heads than PointNet AE.

\begin{table}[]
\centering
\begin{tabular}{l|c|c|c}
                 &  EL &  area ($10^{-4}$) &volume ($10^{-5}$)  \\\hline
PointNet AE (unsupervised) &0.597& 3.508 & 5.251 \\
Ours (unsupervised) &\textbf{0.398}&\textbf{2.752}&\textbf{4.718}\\
\end{tabular}
\caption{We report the mean squared variance of the edge length (EL), per surface area and total shape volume over the interpolations of 100 shape pairs.
We highlight, while both models produce worse results than their supervised equivalents, our method leads to better interpolations.}
\label{tab:interpolation_unsupervised}

\end{table}

\section{Architecture details}
\label{sec_supp:architecture}

\begin{figure}[t]
\begin{center}
   \includegraphics[width=0.7\linewidth]{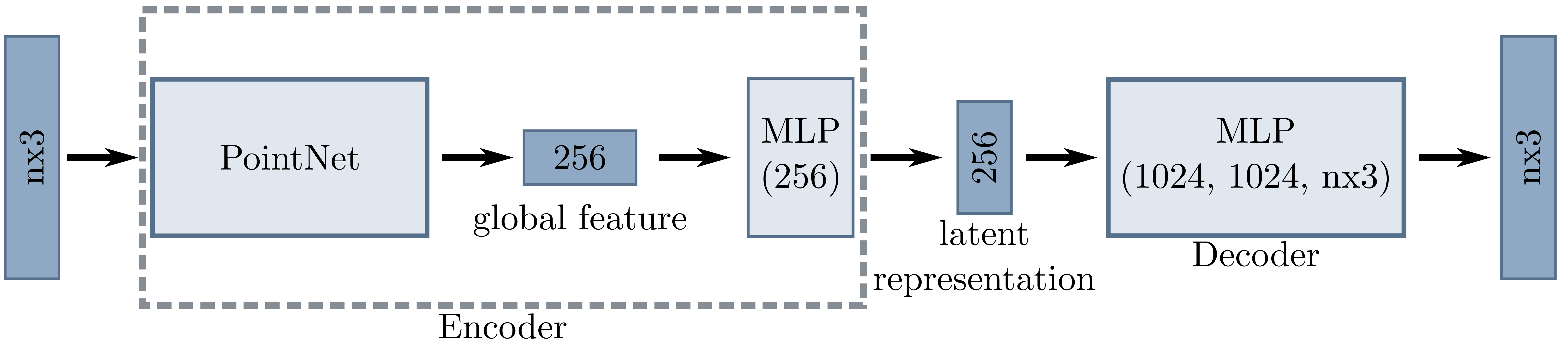}
\end{center}
   \caption{Shape AE architecture.}
\label{fig:shape_ae_archi}
\end{figure}

\begin{figure}[t]
\begin{center}
   \includegraphics[width=0.7\linewidth]{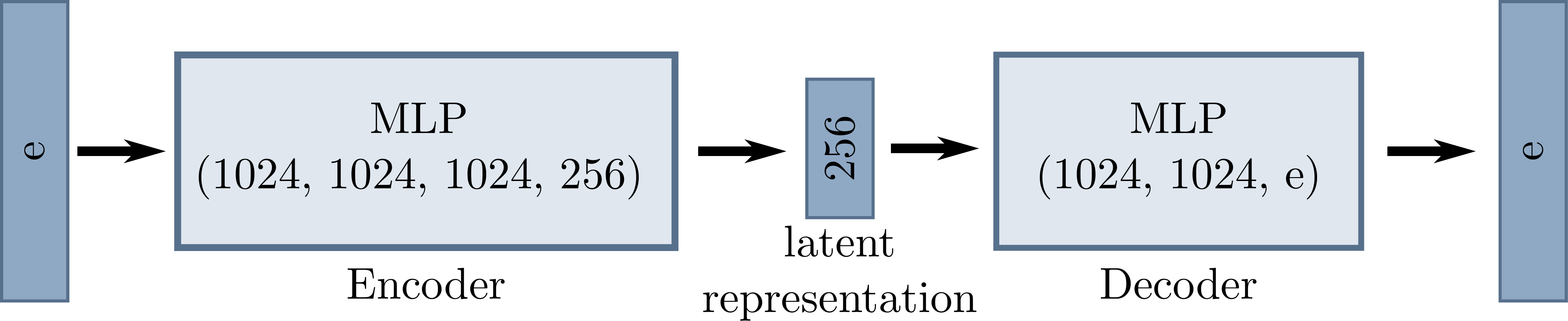}
\end{center}
   \caption{Edge length AE architecture.}
\label{fig:edge_ae_archi}
\end{figure}

\begin{figure}[t]
\begin{center}
   \includegraphics[width=0.5\linewidth]{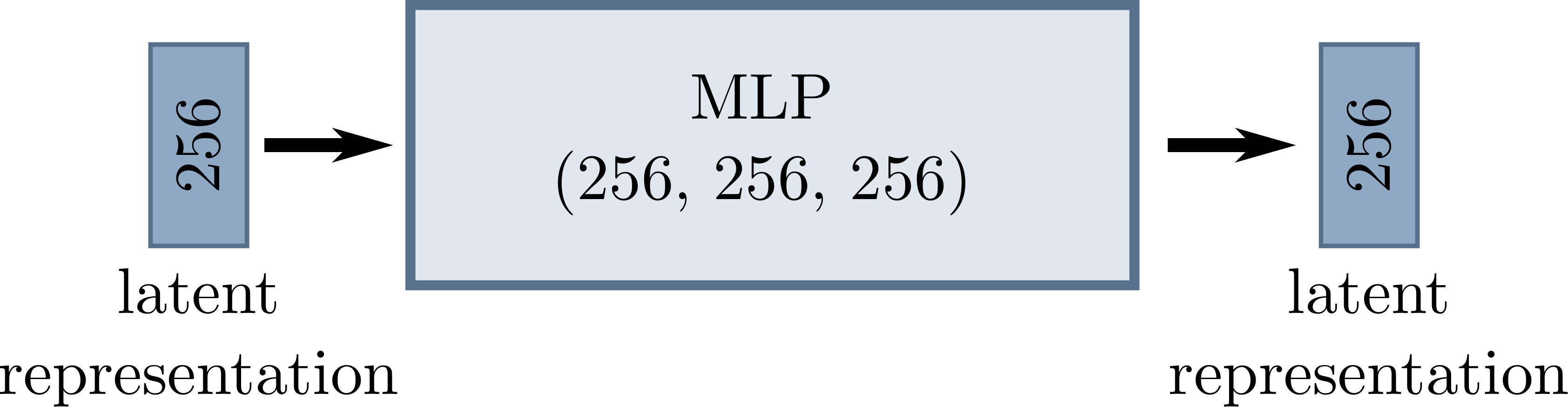}
\end{center}
   \caption{Mapping networks architecture.}
\label{fig:mapping_archi}
\end{figure}

We present the detailed architecture of the shape AE, edge length AE and mapping networks in Figure \ref{fig:shape_ae_archi}, \ref{fig:edge_ae_archi}, \ref{fig:mapping_archi}.

We implemented the presented architectures using Tensorflow and the Adam optimizer for training. Our complete implementation will be released upon acceptance.

\end{document}